\documentclass{llncs}
\usepackage[english]{babel}
\usepackage{amsfonts}
\usepackage{amssymb}
\usepackage{epigraph}
\usepackage[cmex10]{amsmath}

\usepackage{ifpdf}

  \usepackage[pdftex]{graphicx}
  \graphicspath{{../pdf/}{../jpeg/}{./BRS/}}
\usepackage[pdftex,
            pagebackref=true,
            colorlinks=true,
            linkcolor=blue
           ]{hyperref}

\usepackage[tight,footnotesize]{subfigure}
\def\email#1{\href{mailto://#1}{#1}}

\begin{document}
%

\title{Fractally-organized Connectionist Networks: Conjectures and Preliminary Results}
       \author{Vincenzo De Florio}
       \institute{MOSAIC/University of Antwerp and MOSAIC/iMinds research institute\\
Middelheimlaan 1, 2020 Antwerp, Belgium\\
\email{vincenzo.deflorio@gmail.com}}

\maketitle

\begin{abstract}
A strict interpretation of connectionism mandates complex networks of simple components.
The question here is, is this simplicity to be interpreted in absolute terms?
I conjecture that absolute simplicity might not be an essential attribute of connectionism,
and that it may be effectively exchanged with
a requirement for relative simplicity, namely simplicity with respect
to the current organizational level.
In this paper I provide some elements to the analysis of the above question.
In particular I conjecture that fractally organized connectionist networks
may provide a convenient means to achive what Leibniz calls
an ``art of complication'', namely an effective way to encapsulate complexity
and practically extend the applicability of connectionism to domains
such as sociotechnical system modeling and design.
Preliminary evidence to my claim is brought by considering the design
of the software architecture designed for the telemonitoring service
of Flemish project ``Little Sister''.

\end{abstract}


%

\def\PS{\mathcal{P\kern-.2ex S}}
\def\CONTEXT{\hbox{$\mathcal{C}$}}
\def\C{\hbox{$\mathcal{C}$}}
\def\PREC{\prec_{\mathcal{P}}}
\def\PRECA{\prec_{\mathcal{A}}}
\def\PRECMA{\prec_{\mu\kern-.3ex\mathcal{A}}}
\def\PRECMC{\prec_{\mu\kern-.2ex\mathcal{C}}}
\def\PRECMR{\prec_{\mu\kern-.2ex\mathcal{R}}}



\section{Introduction}\label{s:c}
Connectionism---also known as parallel distributed processing (PDP)
and artificial neural networks (ANN)---has been
successfully applied to several problems, including pattern and object recognition,
speaker identification, face processing, image restoration, medical diagnosis, and others~\cite{ConnectApps},
as well as to several cognitive functions~\cite{Thomas2008-THOCMO}.
In connectionism,
\begin{quote}
``Processing is characterized by patterns of
activation across \textbf{simple} processing units connected together into complex networks.
Knowledge is stored in the strength of the connections between units.''~\cite{Thomas2008-THOCMO}
\end{quote}
The accent on simplicity is also present in another
definition of connectionism:
\begin{quote}
``The emergent processes of interconnected networks of \textbf{simple} units''~\cite{Wittek14}.
\end{quote}
Similarly, in~\cite{Rumelhart:1986:GFP:104279.104286}
Rumelhart, Hinton, and McClelland introduce PDP
as a model based on a set of \textbf{small, feature-like} processing units.

I believe it is important to reflect on the
\emph{simplicity\/} requirement expressed in the above definitions.
Regardless of their position and role, the nodes of a connectionist network are intended as
simple parts. This is the case also when the network is organized into a complex hierarchy of layers.
Simplicity pertains to the function of the role but also to the role played by the node, which
is tuneable though statically defined.

My question here is: is this simplicity to be interpreted \emph{in absolute terms}?
If that would be the case, then individual nodes may not represent complex behaviors resulting
from the collective action of aggregations of other nodes.
In this sense, absolute simplicity of the nodes may limit the applicability of connectionism.
How could one easily and comfortably model, e.g., a complex social organization, or a digital ecosystem,
or a biological organism, only by reasoning in terms of simple nodes?
Such an endeavour would be the equivalent to writing a complex software application
with no mechanism to encapsulate complexity (such as software modules, services,
aspects, and components).

The rest of this article is to detail the reasons why my answer to the above question
is ``no''. In fact, my conjecture is that
absolute simplicity might not be an \emph{essential\/} attribute of connectionism,
and that it may be effectively exchanged with
a requirement for \textbf{relative simplicity}, namely simplicity with respect
to the current organizational level.

A second conjecture here is that a convenient hybrid form of connectionism would be that of
\emph{fractally organized connectionist networks\/} (FOCN).
More formally, 
\textbf{fractal connectionism}
would replace the absolute simplicity requirement of ``pure'' connectionism
with the following two properties:
\begin{description}
\item[Fractal Organization:]
FOCN nodes are fractally organized~\cite{Koe67}. This in particular means that nodes
have a dynamic organizational role that depends on the
context and on system-wide organizational rules---the so-called ``canon''. 
In other words, regardless of their level in a fractal hierarchy, the nodes obey the same canon
and switch between, e.g., management and subordinate, or input and output roles,
depending on the situation at hand.
The nodes become thus \emph{organizationally homogeneous}. In fractal organizations
nodes are typically called holons~\cite{Koe67} or fractals.
\item[Increasing Inclusiveness:]
FOCN nodes function as modules that are at the same time
monadic (namely, atomic and indivisible) with respect to the layer they reside in and
composite organizations of parts residing in lower layers~\cite{HT:TE14a}.
Through these ``organizational digits''
\emph{absolute simplicity becomes relative simplicity}.
As nonterminal symbols in context-free grammars, every
node in FOCN is in itself both a network and the ``root'' of that
network.
\end{description}

In what follows I provide some elements towards a discussion
of the benefits of coupling fractal organization with connectionism.
\begin{itemize}
\item In Sect.~\ref{s:L} I first identify in the so-called Art of Complication of Leibniz the
ancestor of ``relative simplicity'' and fractal organization.
\item In Sect.~\ref{s:FSO} I briefly recall the major aspects of fractal social organizations,
an organizational model for sociotechnical systems and cyber-physical societies.
In particular in that section I compare the major differences between PDP and fractal social
organizations. As a result of that comparison, fractal social organizations are
interpreted as a FOCN organizational model.
\item A practical application of said model is the subject of Sect.~\ref{s:LS}:
the web service architecture and middleware developed in the framework
of Flemish project ``Little Sister''.
\item Conclusions and next steps are then drawn in Sect.~\ref{s:end}.
\end{itemize}

\section{Leibniz' Art of Complication}\label{s:L}
\epigraph{%
When the tables of categories of our \emph{art of complication\/} have been formed,
\emph{something greater will emerge}.
For let the first terms, of the combination of which all others consist, be 
designated by signs; these signs will be a kind of alphabet. It will be convenient for the signs 
to be as natural as possible---e.g., for one, a point; for numbers, points; for the relations of 
one entity with another, lines; for the variation of angles and of extremities in lines, kinds of 
relations. If these are correctly and ingeniously established, this universal writing will be as 
easy as it is common, and will be capable of being read without any dictionary; at the same time, 
a fundamental knowledge of all things will be obtained. The whole of such a writing will be made 
of geometrical figures, as it were, and of a kind of pictures---just as the ancient Egyptians 
did, and the Chinese do today. Their pictures, however, are not reduced to a fixed alphabet [\ldots] 
with the result that a tremendous strain on the memory is necessary, which is the contrary of 
what we propose.~\cite{LeibnizParkinson}}
{\textit{Of the art of combination}\\ \textsc{G.~W. von Leibniz}}

As brilliantly discussed in~\cite{HT:TE14a}, hierarchies are a well-known and consolidated concept that pervades
the organization of both our societies and biological systems.
Particularly interesting and relevant to the present discussion are
so-called nested compositional hierarchies, defined in the cited reference as ``a pattern of relationship
among entities based on the
\emph{principle of increasing inclusiveness}, so that entities at one level are composed of
parts at lower levels and are themselves nested within more extensive entities''.
As mentioned in Sect.~\ref{s:c}, increasing inclusiveness (I${}^2$) practically realizes
modularity and relative simplicity by creating matryoshka-like concepts that are at the same time
monadic and composite. The same principle and the same duality may be found in the philosophy
of Leibniz~\cite{DF14c,DBLP:journals/corr/Florio14d}.
The Great One introduces the concept of substances, namely
\begin{quote}
   ``networks of other substances, together with their relationships.
   [\ldots A substance is a] concept-network packaging a quantum of knowledge that becomes
   a new \emph{digit}: a new concept so unitary and
   indivisible as to admit a new pictorial representation---a new and unique [pictogram]''~\cite{DF14c}.
\end{quote}
Leibnitian pictograms were thus an application of the I${}^2$ principle to knowledge representation.
Pictograms of substances are thus at the same time knowledge units and knowledge networks; unique
digits and assemblies of lower level signs and pictograms; which are obtained through some well-formed method of
composition---some compositional \emph{grammar}. Leibniz called the corresponding
\emph{language\/} ``Characteristica Universalis'' (CU): a knowledge representation language
in which any conceptual model would have been expressed and reasoned upon in a mechanical
way. The ``engine'', or algebra, for crunching CU expressions was called by Leibniz
Calculus Ratiocinator. A parser reducing a sentence in a context-free language into a nonterminal symbol
is a natural example of a Calculus Ratiocinator. As mentioned in~\cite{DF14c},
\begin{quote}
``Such 
pictograms represent modules, namely
knowledge components packaging other ancillary knowledge components. In
other words, pictograms are Leibniz's equivalent of Lovelace's
and Turing's tables of instructions; of subroutines in programming languages;
of boxes in a flowchart; of components in component-based software engineering.''
\end{quote}

Interestingly enough, the same principle and ideas were recently re-introduced in
Actor-Network Theory~\cite{Latour99,Latour06}
through the concepts of \emph{punctualization\/} and \emph{blackboxing}.

\section{Connectionism vs. Fractal Social Organizations}\label{s:FSO}
Fractal social organizations (FSO) are a class of socio-technical systems introduced in~\cite{DF13c,DFSB13a}.
FSO may be concisely described as a fractal organization of nodes called
service-oriented communities (SoC's)~\cite{DFC10}.
Such nodes are ``organizationally homogeneous'', meaning that they provide the same,
relatively simple organizational functions regardless of their place in the FSO network.
Each node is a fractal---in the sense specified in Sect.~\ref{s:c}---and may include
other nodes, thus creating a matryoshka-like structure. A special node withing each SoC \emph{punctualizes\/}
the whole SoC. Such node is called representative and is at the same time both a node of the current SoC
and a node of the ``higher-ups''---namely the SoC's that include the current SoC.
Nodes publish information and they offer and require services. Information and service descriptions reach the
representative and are stored in a local registry. The arrival of new information and service descriptions
triggers the execution of response activities, namely guarded actions that are enabled by the availability
of data and roles. Missing roles triggers so-called exceptions: the request is forwarded to the higher-ups
and the missing roles are sought there. Chains of exceptions propagate the request throughout the FSO and result
in the definition of new temporary SoC's whose aim is executing the response activities.
The lifespan of the temporary SoC's is limited to the execution of the activities they are
associated with. Due to the exception mechanism,
the new temporary SoC's may include nodes that belong to several and possibly ``distant'' SoC's.
As such they represent an overlay network that is cast upon the FSO. Because of this I
call them ``social overlay networks'' (SON).

In order to assess the relationship between FSO and PDP, now I briefly review the components
of the PDP model as introduced by
Rumelhart, Hinton, and McClelland in~\cite{Rumelhart:1986:GFP:104279.104286}.
For each component I highlight similarities and ``differentiae''~\cite{Burek04}, namely
specific differences with respect to elements of the FSO model~\cite{DeFPa15a}.

In what follows, uncited quotes are to be assumed from~\cite{Rumelhart:1986:GFP:104279.104286}.

\begin{figure}[h]
\centerline{\includegraphics[width=0.90\textwidth]{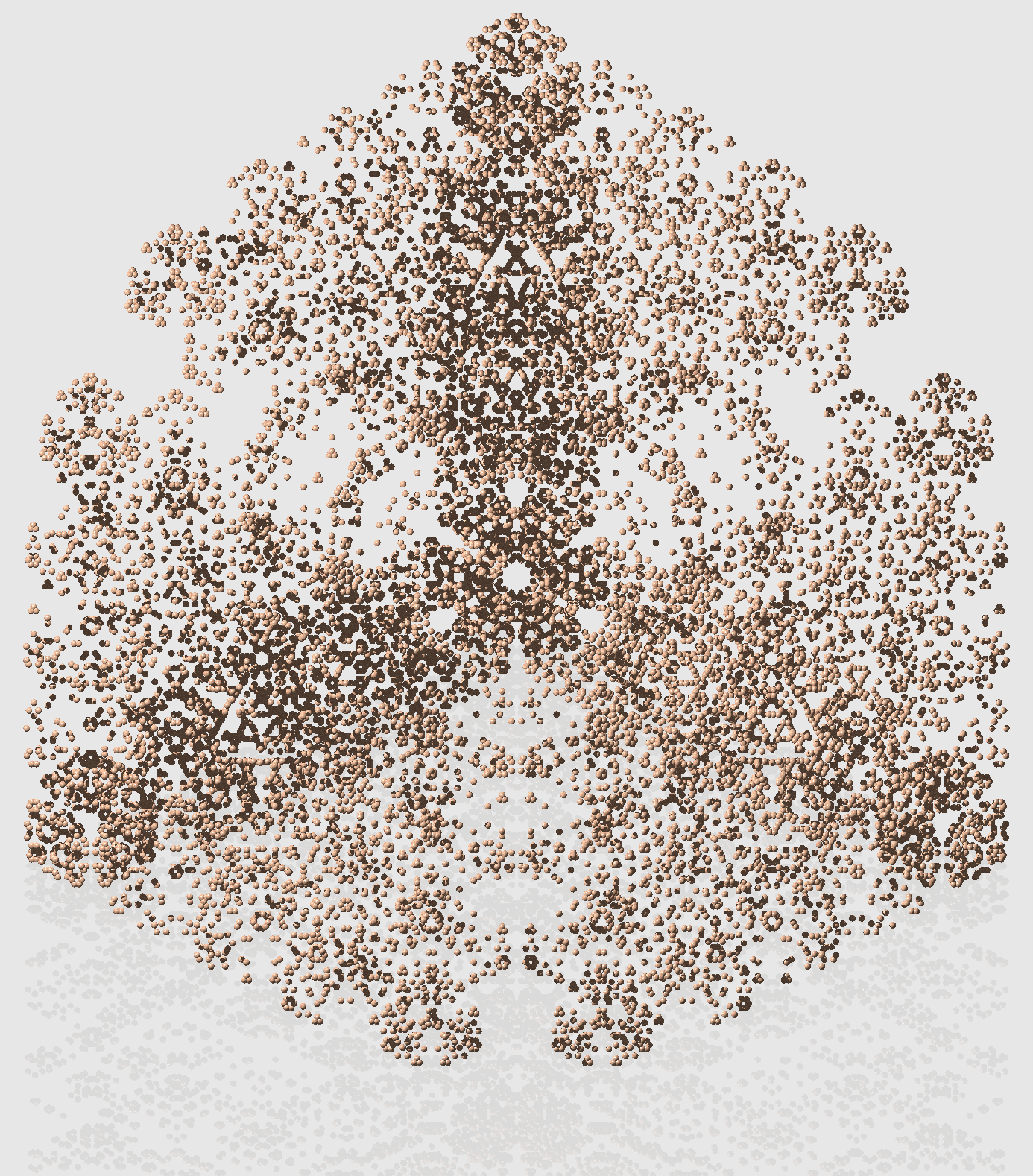}}
\caption{Space of all possible states of activation of an FSO
  with nine agents and six roles. Roles are represented as integers $0,\dots6$. Role 0 is
  played by four agents, all the other roles can be played by one agent only.}\label{f:fso}
\end{figure}

\begin{itemize}
\item  ``A set of processing units''. In PDP these units may represent
``features, letters, words, or concepts'', or they may be ``abstract elements over which
meaningful patterns can be defined''. Conversely, in FSO
those units are \emph{actors}, identified by a set of integers~\cite{DF13c}.
A major difference is that in FSO
actors can be ``small, feature-like entities'' but also complex collections thereof.
Another difference of FSO is given by the presence of a special role---the above mentioned representative.

\item  ``A state of activation''. In PDP this refers to the range of states the processing nodes may assume
over time.
In PDP this range may be discrete or continuous. A simple example is given by range $\{0,1\}$, interpreted
as ``node is inactive'' and ``node is active''.

In FSO the state of activation is simply whether an actor is involved in
an activity and thus is playing a role, or if it is inactive. In~\cite{DF13c} I described
the global state of activation of FSO by means of two dynamic systems,
$L(t)$ and $R$(t), respectively representing all inactive and all active FSO actors at time $t$.
Pictures such as in Fig.~\ref{f:fso}
represent the space of all possible states of activation of an FSO.
Actors can request services or provide services---which corresponds to the input and output units
in~\cite{Rumelhart:1986:GFP:104279.104286}. The visibility of actors is restricted
by the FSO concept of \emph{community}: a set of actors in physical
or logical proximity, for the sake of simplicity interpreted as a \emph{locus\/} (for instance
a room; or a building; or a city, etc.)
Non-visible actors correspond to the hidden units of PDP~\cite{Rumelhart:1986:GFP:104279.104286}.
\item  
The behaviors produced by the activated actors of an FSO correspond to what Rumelhart et al. call as the
``output of the units'' in PDP. In FSO, this behavior is cooperative and is mediated by the representative node.
In the current implementation of our FSO models,  a node's output is equal to the
state of activation. 
In other words, in FSO an actor is currently either totally involved in playing a role or not at all.
Future, more realistic implementations will introduce a percentage of involvement, corresponding
to PDP's unit output. This will make it possible to model involvement of the same actor
in multiple activities.

\item
Nodes of a PDP network are also characterized by a ``pattern of connectivity'', namely the interdependencies among
the nodes. Each PDP node, say node $n$, has a \emph{fan-in\/} and a \emph{fan-out}, respectively meaning
the number of nodes that may have an influence on $n$ or the number of nodes that may be influenced by $n$.
Influence has a \emph{sign}, meaning that the corresponding action may be either excitatory or inhibitory.
Conversely, in FSO I distinguish two phases---construction and reaction. In \emph{construction}, the only pattern
of connectivity is between the nodes of an SoC and the SoC representative. This pattern extends beyond
the originating SoC by means of the mechanism of exception and results in the definition of a new
temporary SoC---the already mentioned SON. Once this is done, \emph{reaction\/} takes place with the enaction
of all the SON agents. Different patterns of activity may emerge at this point, representing
how each SON agent contributes to the emerging collective behavior of the SON.

\item
Another element of the PDP model is the ``rule of propagation'', stating how ``patterns of activities
[propagate] through the network of connectivities'' in response to an input condition.
In FSO, propagation is simply regulated by the
canon, namely the rules of the representative and of the exception~\cite{DeFPa15a}.

\item So-called ``activation rule'' is a function modeling the next state of activation given the current one
and the state of the network. In the current model, FSO activation rules are very simple and dictate that
any request for enrollment to an inactive role is answered positively.
A more realistic implementation should model the propensity and condition of a node to accept a request
for enrollment in an activity. Factors such as the availability of the node, its current output level
(namely, degree of involvement), and even economic considerations such as intervention policies and ``fares''
should be integrated into our current FSO model.

\item Another important component of the PDP model is ``Modifying patterns of connectivity
as a function of experience''. As suggested in our main reference, ``this can involve three types of
modifications:
\begin{itemize}
\item The development of new connections.
\item The loss of existing connections.
\item The modification of the strengths of connections that already exist.''
\end{itemize}
As mentioned above, in FSO we have two types of connections:
\begin{enumerate}
\item ``Institutional'' connections,
represented by relationships between organizationally stable SoC's. An example is
a ``room'' SoC that is stably a part of a ``smart house'' SoC, in turn a stable member
of a ``smart building'' SoC.
\item ``Transitional'' connections,
namely connections between existing SoC's and new SON's.
\end{enumerate}
As I suggested in~\cite{DF13c}, experience may play an important role in FSO too.
By tracking the ``performance'' of individual nodes (as described, e.g., in~\cite{BDFB12,Buys2011})
and individual SON's the structure and processing of an FSO may \emph{evolve}.
In particular, transient SON may be recognized as providing a recurring ``useful'' function,
and could be ``permanentified'' (namely, turned into a new permanent SoC).
An example of this may be that of a so-called ``shadow responder''~\cite{DFSB14}
providing consistently valuable support in the course of a crisis management action. 
Permanentification would mean that the shadow responder---for instance,
a team of citizens assembled spontaneously and providing help and assistance to the
victims of a natural disaster---would be officially or de facto recognized and
integrated in the ``institutional'' response organizations, as suggested
in~\cite{RAND,CARRI3}.

Similarly, SoC that repeatedly fail to provide an effective answer to experienced situations
may cease to make sense and be removed from the system. Reorganizations are a typical
example of cases in which this phenomenon may occur.

The PDP concept of the strength of connection is also both interesting and relevant to the present discussion.
A connection between two nodes may be realized as being ``mutually satisfactory'' (what is
sometimes called as a ``win-win'') and in the long run may strengthen by producing a stable
connection. Mutualistic relationships such as symbiosis and commensalism are typical examples of this
phenomenon. Their role in FSO has been highlighted in~\cite{DBLP:journals/corr/FlorioSB14}.

\item A final element in PDP is the ``representation of the environment''. This is a key
component in the FSO model too, though with a very different interpretation of what an
environment is. In PDP environment is ``a time-varying stochastic function over the space of
input patterns'', while in FSO is is the set of probabilistic distributions representing
the occurrence of the input events. As an example, environment is interpreted as FSO
also as the rate at which new requests for assistance enter the FSO.
\end{itemize}

\section{The Little Sister Software Architecture}\label{s:LS}
\begin{figure}
\centerline{\includegraphics[width=0.90\textwidth]{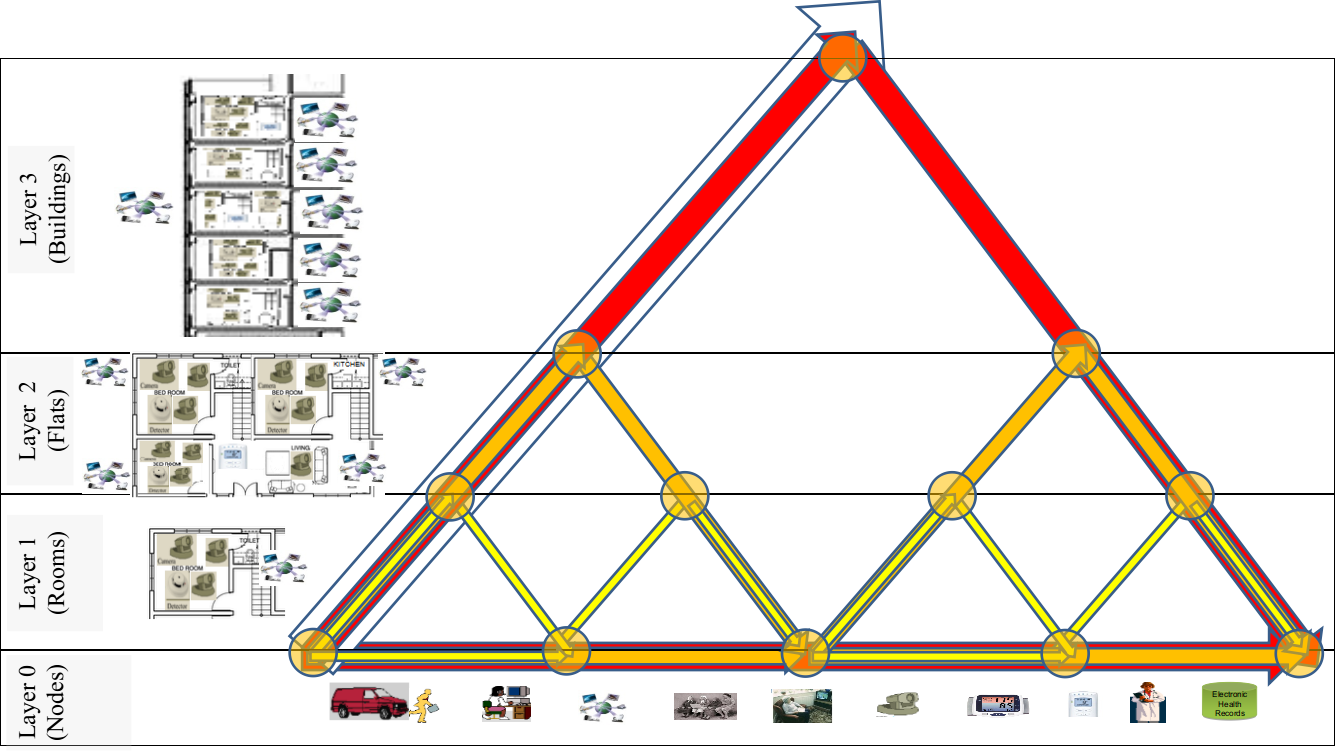}}
\caption{Exemplification of the LS Fractal Social Organization.
}\label{f:lsfso}
\end{figure}

Little Sister (LS) is the name of a Flemish ICON project financed by
the iMinds research institute and the Flemish Government Agency for Innovation by
Science and Technology. The project run in 2013 and 2014 and aimed to deliver a
low-cost telemonitoring~\cite{Meystre05} solution for home care.
Two are the reasons for mentioning LS here:
\begin{figure}
\centerline{\hbox{%
\includegraphics[height=0.2618\textwidth]{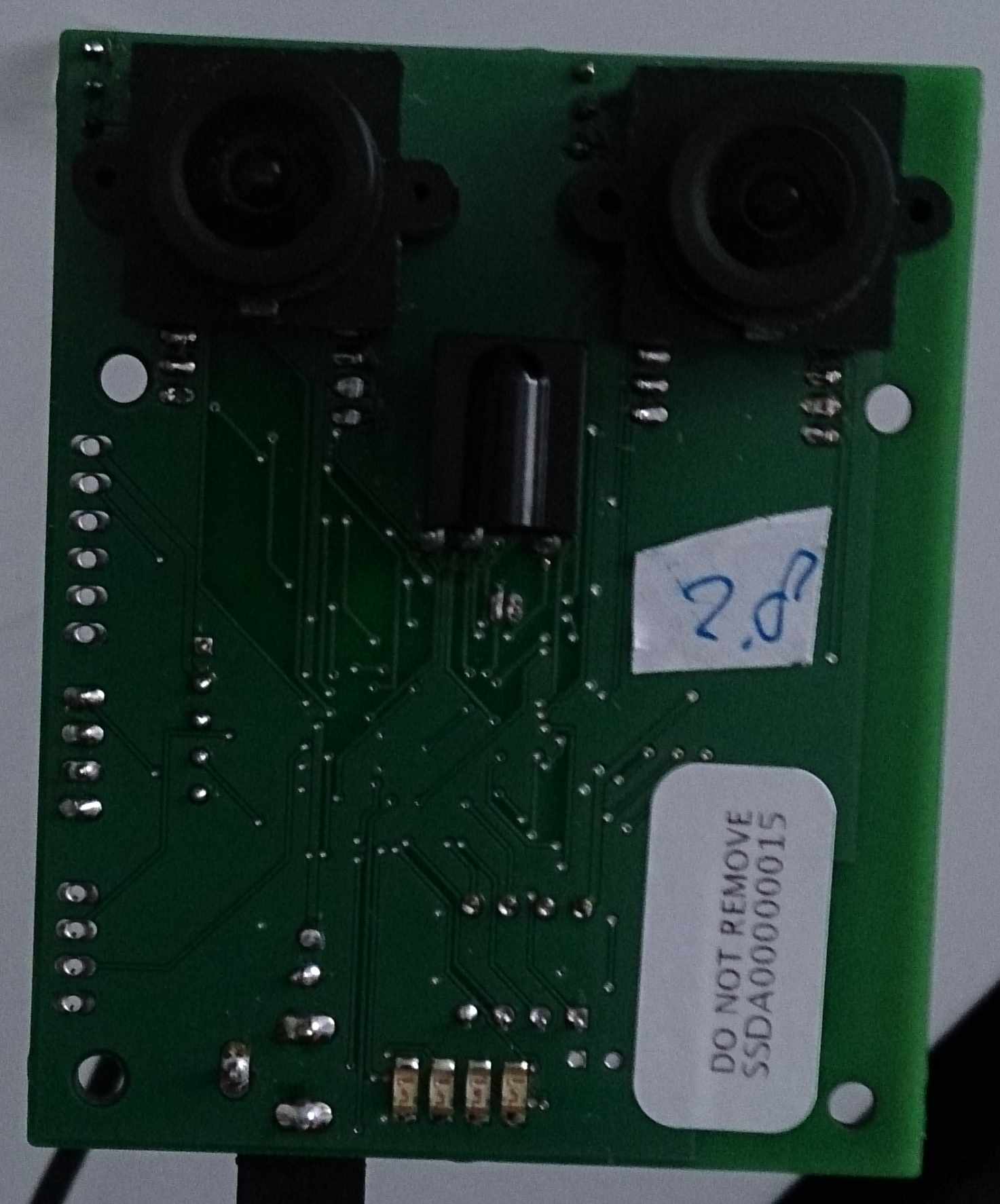}%
\hspace*{32000sp}
\includegraphics[width=0.26\textwidth]{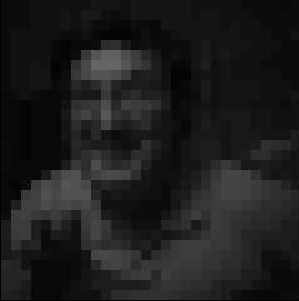}%
}}
\caption{The Little Sister mouse-cam sensor~\cite{IGO} and an exemplary picture taken with it.}\label{f:sensors}
\end{figure}
\begin{enumerate}
\item LS may be considered as a connectionist approach to telemonitoring: in fact in LS
the collective action of an interconnected network
of simple units~\cite{IGO} (battery-powered mouse sensors) replaces the adoption of more powerful and
expensive complex devices (smart cameras; see Fig.~\ref{f:sensors}).
\item The LS software architecture realizes a simplified FSO: a predefined set of SoC's realizes
the structure exemplified in Fig.~\ref{f:lsfso}.
\end{enumerate}

The cornerstone of the LS software architecture is given by web services standards.
As discussed in~\cite{DFSB13a}, 
\begin{quote}
``the LS mouse sensors are individually
wrapped and exposed as manageable web services.  These services are
then structured within a hierarchical federation reflecting the
architectural structure of the building in which they are
deployed~\cite{Oasis1}.  More specifically, the system maintains
dedicated, manageable service groups for each room in the building,
each of which contains references to the web service endpoint of the
underlying sensors (as depicted in layers 0 and 1 in Fig.~\ref{f:lsfso}).  These
`room groups' are then aggregated into service groups
representative of individual housing units. Finally, at the highest
level of the federation, all units pertaining to a specific building
are again exposed as a single resource (layer 3).  All services and
devices situated at layers 0--3 are deployed and placed within the
building and its housing units; all services are exposed as
manageable web services and allow for remote reconfiguration.''
\end{quote}

Absolute simplicity is here traded with modularity and relative simplicity: each ``level'' hosts
nodes that are ``simple'' with respect to the granularity of the action.
Correspondingly, each layer hosts services of increasing complexity, ranging from image to motion processing and
from raw context perception to situation identification~\cite{YeDM12}.
Each SoC is managed by a representative implemented as a module of a middleware.
Said middleware is
based on a fork of Apache MUSE---``a Java-based implementation of the WS-ResourceFramework (WSRF),
WS-BaseNotification (WSN), and WS-DistributedManagement (WSDM) specifications''~\cite{MUSE} on top
of Axis2~\cite{Axis2}, and partially implements the 
WSDM-MOWS specification~\cite{MOWS} (Web Services Distributed Management: Management of Web Services).

It is the LS middleware component in each SoC that manages the FSO canon:
events produced by the local nodes are received by the middleware by means of
a standardized, asynchronous publish-and-subscribe mechanism~\cite{Oasis2}.
The middleware then verifies whether any of the local nodes may respond with 
some actuation logic. If so, the local node is appointed to the management
of the response; otherwise, an exception takes place (see Sect.~\ref{s:FSO}) and
the event is propagated to the higher-up SoC.
Given the fact that, in LS, a predefined population of nodes and services is
available and known beforehand, the selection and exception mechanisms are
simple and have been implemented by annotating events and services
with topic identifiers. In a more general implementation of the FSO model,
selection and exception require semantic description and matching support
as discussed in~\cite{SDB13a}.

\section{Conclusions}\label{s:end}
\epigraph{At a low level of ambition but with a high degree of confidence [General Systems Theory]
 aims to point out similarities in the 
theoretical constructions of different disciplines, where these exist, and to develop theoretical models having 
applicability to at least two different fields of study. At a higher level of ambition, but with perhaps a lower 
degree of confidence it hopes to develop something like a ``spectrum'' of theories---a system of systems which may 
perform the function of a ``gestalt'' in theoretical construction. Such ``gestalts'' in special fields have been of 
great value in directing research towards the gaps which they reveal.
[\ldots]
Similarly a ``system of systems'' might be of value in directing the attention of theorists
toward gaps in theoretical models, and might even be of value in pointing towards methods of filling them.}
{\textit{General Systems Theory---The Skeleton of Science}\\ \textsc{K. Boulding}}

In this work I considered two seemingly unrelated ``gestalts'': connectionism and fractal organization.
By reasoning about them in general and abstract terms, I observed how connectionism could possibly benefit
from the application of I${}^2$, namely the principle of increasing inclusiveness, and interpret
processing nodes' simplicity in relative rather than absolute terms. I have conjectured that, in so doing,
connectionism would further extend its applicability and expressiveness. I called
fractally-organized connectionist networks the resulting hybrid formulation.
I then introduced a model of fractal organization called FSO and I compared the key elements
of parallel distributed processing with corresponding assumptions and strategies in FSO.
As a practical example of the hybrid model I discussed the software architecture of 
Flemish project ``Little Sister''---a web services-based implementation
of a ``fractally connectionist'' system.
As observed by Boulding~\cite{Bou56}, the above discussion put to the foreground a number
of oversimplifications in the current FSO model.
As a consequence,
our future research shall be
\emph{directed towards the gaps that the above discussion helped revealing},
in particular extending the FSO model with more complete and general elements
of the connectionist models.



\subsection*{Acknowledgment}
This work was partially supported by
iMinds---Interdisciplinary institute for Technology, a research institute
funded by the Flemish Government---as well as by the Flemish Government Agency for Innovation by
Science and Technology (IWT).
The iMinds Little Sister project is a project co-funded by iMinds with project support of IWT
(Interdisciplinary institute for Technology)
Partners
involved in the project are
Universiteit Antwerpen,
Vrije Universiteit Brussel,
Universiteit Gent,
Xetal,
Niko Projects,
JF Oceans BVBA,
SBD NV,
and Christelijke Mutualiteit vzw.

\bibliographystyle{splncs}
\bibliography{/refs/thesis}

\begin{thebibliography}{10}

\bibitem{ConnectApps}
Mira, J., Prieto, A., eds.:
\newblock Bio-Inspired Applications of Connectionism.
\newblock In Mira, J., Prieto, A., eds.: Proc. of the 6th Int.l Work-Conference
  on Artificial and Natural Neural Networks (IWANN 2001). Volume 2085 of
  Lecture Notes in Computer Science., Granada, Spain, Springer, Berlin /
  Heidelberg (2001)

\bibitem{Thomas2008-THOCMO}
Thomas, M.S., McClelland, J.L.:
\newblock Connectionist models of cognition.
\newblock In Sun, R., ed.: The Cambridge Handbook of Computational Psychology.
\newblock Cambridge University Press (2008)

\bibitem{Wittek14}
Wittek, P.:
\newblock Quantum Machine Learning: What Quantum Computing Means to Data
  Mining.
\newblock Academic Press, San Diego, CA (2014)

\bibitem{Rumelhart:1986:GFP:104279.104286}
Rumelhart, D.E., Hinton, G.E., McClelland, J.L.:
\newblock A general framework for parallel distributed processing.
\newblock In Rumelhart, D.E., McClelland, J.L., PDP Research~Group, C., eds.:
  Parallel Distributed Processing: Explorations in the Microstructure of
  Cognition, Vol. 1.
\newblock MIT Press, Cambridge, MA, USA (1986)  45--76

\bibitem{Koe67}
Koestler, A.:
\newblock The Ghost in the Machine.
\newblock Macmillan (1967)

\bibitem{HT:TE14a}
T{\"e}mkin, I., Eldredge, N.:
\newblock Networks and hierarchies: Approaching complexity in evolutionary
  theory.
\newblock In Serrelli, E., Gontier, N., eds.: Macroevolution: Explanation,
  Interpretation, Evidence.
\newblock Springer (2014)

\bibitem{LeibnizParkinson}
Leibniz, G.W.:
\newblock Of the art of combination (1666).
\newblock In: Leibniz: Logical Papers. Clarendon Press, Oxford (1966) {E}nglish
  translation of a portion of Chapter~11 by G.~H.~R.~Parkinson.

\bibitem{DF14c}
{De Florio}, V.:
\newblock Behavior, organization, substance: Three gestalts of general systems
  theory.
\newblock In: Proceedings of the IEEE 2014 Conference on Norbert Wiener in the
  21st Century, IEEE (2014)

\bibitem{DBLP:journals/corr/Florio14d}
{De Florio}, V.:
\newblock Systems, resilience, and organization: Analogies and points of
  contact with hierarchy theory.
\newblock CoRR \textbf{abs/1411.0092} (2014)

\bibitem{Latour99}
Latour, B.:
\newblock Pandora's hope: essays on the reality of science studies.
\newblock Harvard University Press, Cambridge, MA (1999)

\bibitem{Latour06}
Latour, B.:
\newblock On actor-network theory. a few clarifications plus more than a few
  complications.
\newblock Soziale Welt \textbf{47} (1996)  369--381

\bibitem{DF13c}
{De~Florio}, V., Bakhouya, M., Coronato, A., {Di Marzo Serugendo}, G.:
\newblock Models and concepts for socio-technical complex systems: Towards
  fractal social organizations.
\newblock Systems Research and Behavioral Science \textbf{30}(6) (2013)

\bibitem{DFSB13a}
{De Florio}, V., Sun, H., Buys, J., Blondia, C.:
\newblock On the impact of fractal organization on the performance of
  socio-technical systems.
\newblock In: Proceedings of the 2013 International Workshop on Intelligent
  Techniques for Ubiquitous Systems (ITUS 2013), Vietri sul Mare, Italy, IEEE
  (2013)

\bibitem{DFC10}
{De~Florio}, V., Blondia, C.:
\newblock Service-oriented communities: Visions and contributions towards
  social organizations.
\newblock In Meersman, R., Dillon, T., Herrero, P., eds.: On the Move to
  Meaningful Internet Systems: OTM 2010 Workshops. Volume 6428 of Lecture Notes
  in Computer Science.
\newblock Springer Berlin / Heidelberg (2010)  319--328
  10.1007/978-3-642-16961-8\_51.

\bibitem{Burek04}
Burek, P.:
\newblock Adoption of the classical theory of definition to ontology modeling.
\newblock In Bussler, C., Fensel, D., eds.: Proc. of the 11th Int.l Conf. on
  Artificial Intelligence: Methodology, Systems, and Applications (AIMSA 2004),
  LNAI 3192, Springer (2004)

\bibitem{DeFPa15a}
{De Florio}, V., {Pajaziti}, A.:
\newblock {How Resilient Are Our Societies? Analyses, Models, and Preliminary
  Results}.
\newblock ArXiv e-prints (2015)

\bibitem{BDFB12}
Buys, J., {De Florio}, V., Blondia, C.:
\newblock Towards parsimonious resource allocation in context-aware n-version
  programming.
\newblock In: Proceedings of the 7th IET System Safety Conference, The
  Institute of Engineering and Technology (2012)

\bibitem{Buys2011}
Buys, J., {De Florio}, V., Blondia, C.:
\newblock Towards context-aware adaptive fault tolerance in {SOA} applications.
\newblock In: Proceedings of the 5th ACM International Conference on
  Distributed Event-Based Systems (DEBS), Association for Computing Machinery,
  Inc. (ACM) (2011)  63--74

\bibitem{DFSB14}
{De~Florio}, V., Sun, H., Blondia, C.:
\newblock Community resilience engineering: Reflections and preliminary
  contributions.
\newblock In Majzik, I., Vieira, M., eds.: Software Engineering for Resilient
  Systems. Volume 8785 of Lecture Notes in Computer Science.
\newblock Springer International Publishing (2014)  1--8

\bibitem{RAND}
{RAND}:
\newblock Community resilience (2014) Available at URL
  \url{http://www.rand.org/topics/community-resilience.html}.

\bibitem{CARRI3}
Colten, C.E., Kates, R.W., Laska, S.B.:
\newblock Community resilience: Lessons from new orleans and hurricane katrina.
\newblock Technical Report~3, Community and Regional Resilience Institute
  (CARRI) (2008)

\bibitem{DBLP:journals/corr/FlorioSB14}
{De Florio}, V., Sun, H., Bakhouya, M.:
\newblock Mutualistic relationships in service-oriented communities and fractal
  social organizations.
\newblock CoRR \textbf{abs/1408.7016} (2014)

\bibitem{Meystre05}
Meystre, S.:
\newblock The current state of telemonitoring: a comment on the literature.
\newblock Telemed J E Health \textbf{11}(1) (2005)  63--69

\bibitem{IGO}
Anonymous:
\newblock Introducing the {S}ilicam {IGO} (2013)

\bibitem{Oasis1}
OASIS:
\newblock Web services service group 1.2 standard.
\newblock Technical report, OASIS (2006)

\bibitem{YeDM12}
Ye, J., Dobson, S., McKeever, S.:
\newblock Situation identification techniques in pervasive computing: A review.
\newblock Pervasive and Mobile Computing \textbf{8}(1) (2012)  36--66

\bibitem{MUSE}
Anonymous:
\newblock Apache {Muse}---a {Java}-based implementation of {WSRF} 1.2, {WSN}
  1.3, and {WSDM} 1.1 (2010) Retrieved on February 11, 2010 from
  \textsf{ws.apache.org/muse}.

\bibitem{Axis2}
Anonymous:
\newblock Apache {Axis2/Java}---next generation web services (2010) Retrieved
  on February 11, 2010 from \textsf{ws.apache.org/axis2}.

\bibitem{MOWS}
Anonymous:
\newblock Web services distributed management: Management of web services
  ({WSDM-MOWS}) 1.0 (2004) Retrieved on May 17, 2015 from
  \url{http://xml.coverpages.org/WSDM-CD-10971-MOWS10.pdf}.

\bibitem{Oasis2}
OASIS:
\newblock Web services base notification 1.3 standard.
\newblock Technical report, OASIS (2006)

\bibitem{SDB13a}
Sun, H., {De~Florio}, V., Blondia, C.:
\newblock Implementing a role based mutual assistance community with semantic
  service description and matching.
\newblock In: Proc. of the Int.l Conference on Management of Emergent Digital
  EcoSystems (MEDES). (2013)

\bibitem{Bou56}
Boulding, K.:
\newblock General systems theory---the skeleton of science.
\newblock Management Science \textbf{2}(3) (1956)

\end{thebibliography}

\end{document}